\documentclass[conference]{IEEEtran}

\usepackage{times}
\usepackage[numbers]{natbib}
\usepackage{multicol}
\usepackage[bookmarks=true]{hyperref}
\usepackage{amsmath,amssymb,amsfonts}
\usepackage{algorithmic}
\usepackage{graphicx}
\usepackage{textcomp}
\usepackage{xcolor}
\usepackage{booktabs}
\usepackage{tabularx}
\usepackage{threeparttable}
\usepackage{multirow}
\usepackage[most]{tcolorbox}
\tcbuselibrary{listings,breakable,skins}

\newtcblisting{FullWidthPromptBox}[2][]{%
  enhanced,
  breakable,
  float*=t,            %
  width=\textwidth,
  colback=gray!6,      %
  colframe=black,
  colbacktitle=gray!18,
  coltitle=black,
  fonttitle=\bfseries,
  boxrule=0.9pt,
  arc=3mm,             %
  outer arc=3mm,
  drop shadow,
  left=3mm,right=3mm,top=2mm,bottom=2mm,
  listing engine=listings,
  listing only,
  title={#2},
  listing options={
    basicstyle=\ttfamily\footnotesize,
    breaklines=true,
    breakatwhitespace=false,
    columns=fullflexible,
    keepspaces=true,
    showstringspaces=false
  },
  #1
}

\newcommand{\ourmethod}[1]{\textcolor{black}{HALO}}

\begin{document}

\title{Memory Retrieval in Visuomotor Policies for Long-Horizon Robot Control}

\IEEEoverridecommandlockouts
\author{
Rutav Shah$^{1}$, 
Yisu Li$^{*}$, 
Femi Bello$^{*}$, 
Yuke Zhu, 
Roberto Mart{\'i}n-Mart{\'i}n \\
\vspace{-8pt}
\thanks{$^1$ Corresponding author: rutavms@utexas.edu, $^{*}$ Equal contribution} \\
The University of Texas at Austin
\vspace{-5pt}
}

\maketitle

\begin{abstract}
General-purpose robots operating in partially observable environments, such as homes, require memory to support autonomy.
They must recall diverse information from the past, such as where objects were placed, which tasks a human partner has completed, and when an appliance was turned on, to accomplish a wide range of tasks.
Achieving this versatility requires a general memory retrieval mechanism that works across diverse settings.
However, hand-designed and heuristic-based methods rely on task-specific assumptions that limit their generalizability.
Transformer architectures that use attention over long contexts for memory retrieval provide a promising alternative, as they learn retrieval from data rather than relying on task-specific assumptions.
However, directly incorporating long-context transformer architecture into imitation learning from offline data introduces two key challenges: (1) the policy may learn spurious correlations between the past information and predicted actions, and (2) errors accumulate over time in the memory due to prediction inaccuracies and their compounding interactions with the environment, leading to model drift and cascading failures in long-horizon control.
To address both challenges, we introduce \ourmethod{}, a visuomotor policy with an attention-based memory retrieval mechanism for long-horizon control.
First, to suppress spurious correlations, \ourmethod{} leverages vision-language model (VLM) priors to steer retrieval toward task-relevant information. Concretely, it generates task-relevant, memory-dependent question--answer pairs from demonstration trajectories and trains the policy jointly with a video question--answering objective, transferring VLM priors to the visuomotor policy.
Second, to reduce the impact of accumulated errors in memory during closed-loop control, \ourmethod{} uses sparse attention that restricts retrieval to only the most relevant parts of the history.
Together, these components enable more reliable long-horizon control by guiding the policy to retrieve task-relevant information from up to eight minutes of past experience.
Project website: \href{https://robin-lab.cs.utexas.edu/HALO}{robin-lab.cs.utexas.edu/HALO}
\end{abstract}

\IEEEpeerreviewmaketitle

\section{Introduction}
\begin{figure}[htbp]
    \centering
    \includegraphics[width=0.5\textwidth, trim=0 0.5cm 0 0, clip]{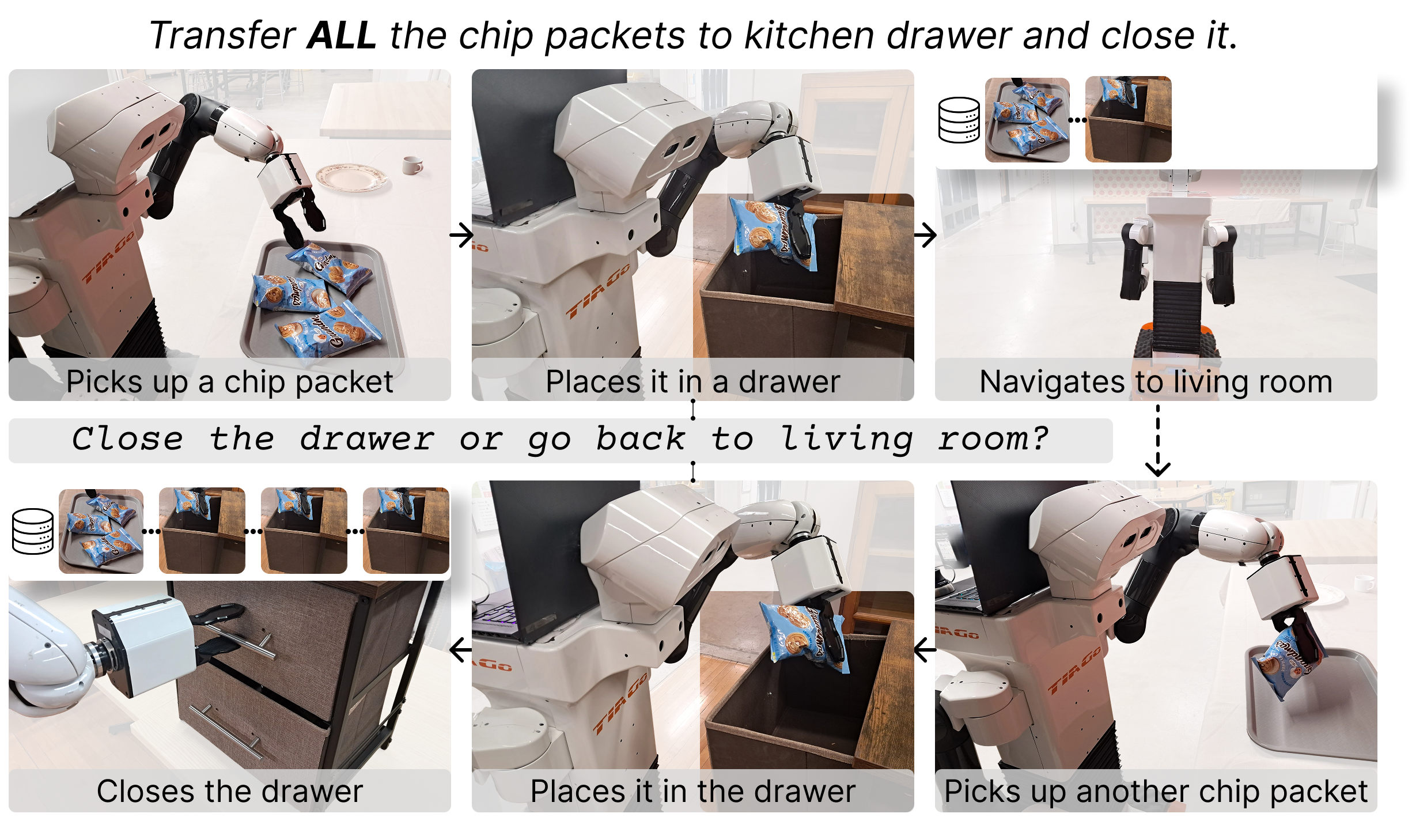}
    \vspace{-20pt}
    \caption{\textbf{Memory retrieval in visuomotor policy learning.} Long-horizon household tasks require robots to act on information no longer present in the current sensory input. Under partial observability, the policy must retrieve relevant information from past observations stored in memory to predict correct low-level actions. The type of information needed can change across task stages, motivating a general, learned retrieval mechanism.}
    \label{fig:pull_figure}
    \vspace{-10pt}
\end{figure}

General-purpose robots operating in household environments must perform long-horizon manipulation tasks that require recalling information no longer present in current sensory input.
The robot may need to remember where an object was placed earlier, when an appliance was turned on, how many pieces of bread were placed in the oven, or which tasks have already been completed (Fig.~\ref{fig:pull_figure}).
Such requirements arise frequently in everyday household activities, where successfully completing the task depends on recalling information over extended time horizons.
As a result, purely reactive visuomotor policies will fail on these tasks, motivating the need for \textbf{memory retrieval} mechanisms that can leverage previously gathered information~\cite{tulving1972episodic,atkinson1968human}.

The information that robots must recall from history is diverse.
It may include spatial information such as object locations, relational information between objects, numerical quantities, temporal events, or event ordering.
Approaches based on hand-designed memory representations or modality-specific retrieval rules, therefore, struggle to scale across the wide range of manipulation tasks required of general-purpose robots~\cite{guadarrama2014open, clipnav, sam2act}.
This motivates the development of a \textit{general} memory retrieval mechanism that can be learned end-to-end, rather than tailored to individual tasks or modalities~\cite{sukhbaatar2015end,graves2014neural,graves2016hybrid,lstm}.

Among candidate mechanisms for general memory retrieval, associative retrieval based on learned query-key similarity, such as attention in transformers, has emerged as a promising mechanism~\cite{carpenter1989neural,vaswani2017attention}.
By learning both queries and keys from data, associative retrieval avoids modality-specific design, and has demonstrated strong empirical performance in domains such as language processing~\cite{floridi2020gpt}.
However, directly applying attention-based memory retrieval to long-horizon robotic imitation learning via offline data exposes two fundamental challenges.
First, policies may exploit spurious correlations between past observations and current actions, \textit{i.e.}, attending to information that correlates with expert behavior in the training data but is not causally relevant to task execution.
\begin{figure}[htbp]
    \centering
    \includegraphics[width=0.5\textwidth]{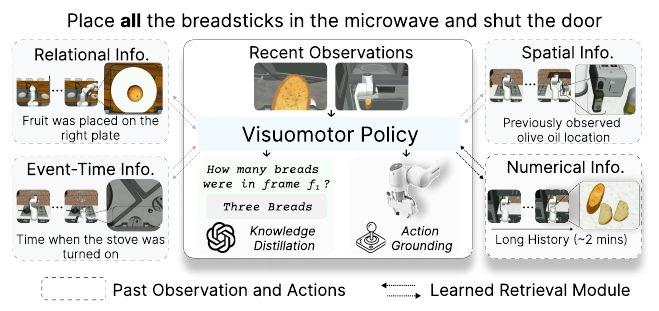}
    \vspace{-20pt}
    \caption{\ourmethod{} learns to retrieve diverse forms of task-relevant information from history, guided by priors distilled from vision-language models.}
    \label{fig:overview}
    \vspace{-10pt}
\end{figure}

Such correlations often fail to hold at test time, leading to poor performance~\cite{de2019causal,wen2020fighting}.
Second, in offline imitation learning, small prediction errors compound during closed-loop online execution~\cite{ross2011reduction,spencer2021feedback}. 
Conditioning on large amounts of past observations can amplify this effect, as the policy repeatedly attends to noisy information, leading to drift and cascading failures over long horizons.
Together, these challenges can limit the effective use of memory in imitation learning, a crucial capability for partially observable long-horizon tasks.

To address these challenges, we propose \ourmethod{}: \textbf{H}istory-\textbf{A}ware visuomotor policy for \textbf{LO}ng-horizon robotic imitation learning.
\ourmethod{} enables end-to-end execution of tasks spanning up to \textit{eight minutes}, requiring memory retrieval across the full duration.
Our key insight is that while attention provides a general retrieval mechanism, it must be guided towards task-relevant information and constrained to avoid conditioning on irrelevant or noisy history (Fig.~\ref{fig:overview}).

To mitigate spurious correlations, we build on the observation that vision-language models (VLMs) encode rich priors about which information from past observations is relevant for a task~\cite{bubeck2023sparks}.
Although these priors are not grounded in action and are therefore insufficient on their own for control, they provide useful signals for identifying task-relevant information in history.
In \ourmethod{}, we distill these priors into the policy through automatically generated video question–answer (VQA) supervision.
To answer these questions correctly, the retrieval module must attend to specific parts of the history where task-relevant event(s) occurred, providing direct supervision on what should be retrieved from memory.
The policy is co-trained to both imitate expert actions and answer questions about past observations. The VQA biases memory retrieval towards task-relevant information, whereas the action prediction objective may still access distinct information needed for low-level control.
This contrasts with prior work that incorporates VLM priors through text-based summaries~\cite{anwar2025remembr,torne2026mem}, which may omit details needed for control.

To address error accumulation over long horizons, \ourmethod{} further restricts the policy to condition only on a sparse subset of stored information.
Concretely, the policy retrieves and processes only the top-$k$ most informative observations or actions via learned query-key matching~\cite{child2019generating}.
Unlike prior works that rely on human priors, such as hand-designed rules~\cite{memer} or object-centric representations~\cite{sam2act}, this sparsification is learned end-to-end, potentially offering a more scalable alternative.
By limiting the amount of noisy information in decision-making,~\ourmethod{} reduces closed-loop drift and mitigates cascading failures during execution.

We evaluate our approach on \textsc{ReMemBench}~\cite{anonymous2026scaling}, a benchmark for long-horizon manipulation tasks that require diverse forms of memory, as well as on five real-world tasks across two robotic platforms, including a fixed-base manipulator and a mobile manipulator.
Across these settings, we show that VQA-induced task priors provide a general solution, improving absolute task success by $7\%$ across diverse tasks and outperforming prior methods based on hand-designed rules ($-12\%$) and commonly used task-specific features ($-21\%$).
We further find that while vision-language models encode informative priors about task-relevant information, policies that use these priors alone achieve only $18\%$ absolute success, compared to $41\%$ for~\ourmethod{}, which combines these priors with action-grounded retrieval.
In addition, we demonstrate that top-$k$ attention sparsification, which reduces model drift, improves absolute performance by $9\%$.
Together, these results highlight the importance of learning to retrieve task-relevant information from memory and grounding retrieval in action prediction for long-horizon control.

\section{Related Work}

\subsection{Imitation Learning under Partial Observability}
Standard imitation learning assumes Markovian observations, causing reactive visuomotor policies to fail under partial observability. Belief-state space methods address this issue by explicitly modeling task state but require task-specific representations and update rules~\cite{aastrom1965optimal}.
Recent work instead incorporates explicit memory into policy.
Some methods rely on object-centric representations to reduce the amount of information from visual inputs, but often limit retrieval to spatial information~\cite{sam2act}. 
Others selectively retrieve past observations using similarity in learned semantic embeddings~\cite{clipnav,okrobot}, or rely on hand-designed rules for selectively storing past observations~\cite{memer}.
Such assumptions restrict their applicability to settings where task-relevant information is well captured by a fixed set of representations or retrieval rules, and may not scale to more diverse long-horizon tasks.

Several approaches learn compact representations of history to support long-horizon reasoning~\cite{fang2019scene, memoryvla}. While effective, learned compression can discard granular information, such as low-level action sequences taken by the robot. These limitations motivate memory retrieval mechanisms that can flexibly access diverse task-relevant information from history without relying on task-specific designs.

\subsection{Knowledge Distillation from Foundation Models}
Vision-language foundation models encode rich knowledge about scenes, objects, and activities from large-scale multimodal data and have been widely used in robotics for perception, planning, affordance learning, and semantic reasoning~\cite{shridhar2022cliport,song2023llm,ahn2022can,gu2023rt,nasiriany2025rt,shah2025bumble}.
Prior work has also distilled knowledge from foundation models into robotic systems by improving parametric memory~\cite{zhou2025chatvla,driess2025knowledge}.
In contrast, relatively little work has explored using foundation models to guide memory retrieval in visuomotor imitation learning, \textit{i.e.}, retrieving information from non-parametric memory~\cite{lewis2020retrieval}.
The non-parametric memory allows capturing episodic details, such as the location of the fruit or the time the stove was turned on. 
Our work builds on these efforts by distilling task-relevant priors from vision-language models to supervise learned memory retrieval, while grounding the policy through action imitation.

\begin{figure*}[t]
    \centering
    \includegraphics[width=\textwidth]{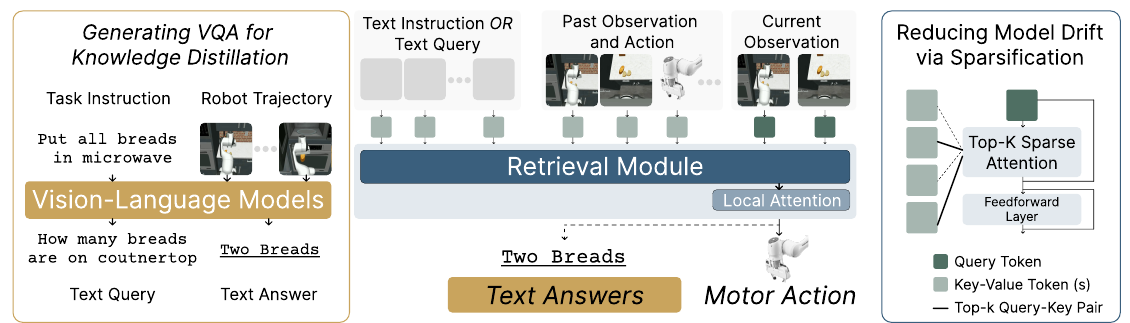}
    \vspace{-20pt}
    \caption{\textbf{Overview.} \ourmethod{} learns a visuomotor policy that retrieves information from the past observations and actions to predict low-level robot actions (middle), guided by priors from vision-language models to retrieve task-relevant information through VQA (left) and uses sparse attention to reduce model drift (right).}
    \label{fig:method_overview}
    \vspace{-10pt}
\end{figure*}

\subsection{Memory Mechanisms for Long-Context Reasoning}
Modeling long-range temporal dependencies is a central challenge in sequential decision making. 
Recurrent and state-space models maintain memory through compressed latent states but struggle with selective recall and credit assignment over long horizons~\cite{lstm, mamba, bengio1993credit}.
Transformer-based models address these limitations by using attention to directly access past inputs~\cite{vaswani2017attention}.
However, dense attention over long contexts can cause irrelevant or noisy information to increasingly influence predictions~\cite{hong2025context}, an issue that is especially pronounced in robotics due to prediction errors and their compounding effects during closed-loop execution.

To regulate information flow in long-context models, prior work has proposed combining attention with recurrence~\cite{dai2019transformer}, a gating mechanism~\cite{gatedattention}, and inducing sparsity in the transformer architecture~\cite{zaheer2020big}.
While these mechanisms have been extensively studied in language and vision, their effectiveness in long-horizon closed-loop visuomotor control has received less attention.
In this work, we find that simple top-$k$ attention sparsification is effective for reducing the influence of accumulated error in model prediction.

\section{\ourmethod{}}
\subsection{Problem Formulation and Overview}
\noindent
\textbf{Problem formulation.}
In this work, we address \emph{long-horizon robotic imitation learning under partial observability}, where successful task execution requires recalling information no longer present in the current sensory input.
At each time step $t$, the robot receives an observation $o_t$ (\textit{e.g.}, RGB images and proprioception) and executes a low-level action $a_t$.
Let
\[
\tau_t = \{(o_1, a_1), \ldots, (o_{t-1}, a_{t-1}), o_t\}
\]
denote the interaction history up to time $t$.
Additionally, we are given a natural-language task instruction $l$ (\textit{e.g.}, ``put all the bread in the microwave and close the door''), which specifies the high-level goal and is provided to the policy at every time step.
We are given a dataset of expert demonstrations
\[
\mathcal{D} = \{\tau^{(i)}\}_{i=1}^N,
\]
where we collect each trajectory
\[
\tau^{(i)} = \{(o^{(i)}_1, a^{(i)}_1), \ldots, (o^{(i)}_{T_i}, a^{(i)}_{T_i})\}
\]
by teleoperating the robot.
Our aim is to learn a closed-loop visuomotor policy
\[
\pi_\theta(a_t \mid \tau_t, l),
\]
that imitates the expert and performs long-horizon tasks by retrieving and processing relevant information from history.

\vspace{0.5em}
\noindent
\textbf{Model architecture overview.}
We parameterize the visuomotor policy $\pi_\theta(a_t \mid \tau_t, l)$ with three main components: 
(i) modality-specific encoders consisting of an observation encoder $g_\theta^{\text{obs}}$ and an action encoder $g_\theta^{\text{act}}$, which map heterogeneous inputs (\textit{e.g.}, RGB images, proprioception, and past actions) into a common latent space,
(ii) a policy backbone $f_\theta$ that performs memory retrieval over the encoded history and processes the information, and
(iii) a task-specific head $h_\theta^{act}$ and $h_\theta^{text}$ that predict low-level control commands and text answers, respectively.

At time step $t$, the current observation is encoded as $x_t = g_\theta^{\text{obs}}(o_t)$ and past actions as $e_i = g_\theta^{\text{act}}(a_i)$ for $i < t$.
The encoded history is
\[
\mathcal{M}_t = \{(x_1, e_1), \ldots, (x_{t-1}, e_{t-1})\}.
\]
Given $\mathcal{M}_t$, the current embedding $x_t$, and the task instruction $l$, the transformer-based policy backbone $f_\theta$ produces a latent state
\[
z_t = f_\theta(\mathcal{M}_t, x_t, l),
\]
This latent state is passed to two prediction heads: an action head $h_\theta^{\text{act}}(z_t)$ for control and a text prediction head $h_\theta^{\text{text}}(z_t)$ for auxiliary supervision.

\vspace{0.5em}
\noindent
\textbf{Method overview.}
A common approach for retrieving information from history is to equip the policy backbone ($f_\theta$) with an \emph{associative memory retrieval mechanism}, typically implemented via learned query-key similarity (\textit{i.e.}, attention)~\cite{vaswani2017attention}.
Such mechanisms provide a general, modality-agnostic way to retrieve information from history, avoiding task-specific designs.

Concretely, the policy backbone $f_\theta$ implements retrieval using a query-key-value formulation.
At each time step $t$, a query vector is computed from the current context,
\[
q_t = f_q(x_t, l),
\]
while each memory element $m_i \in \mathcal{M}_t$ is mapped to a key-value pair,
\[
k_i = f_k(m_i), \qquad v_i = f_v(m_i),
\]
where $f_q$, $f_k$, and $f_v$ are learned functions parameterized by $\theta_q \subset \theta$, $\theta_k \subset \theta$, and $\theta_v \subset \theta$, respectively.
Retrieval is performed by computing similarities between $q_t$ and $\{k_i\}$ and aggregating the corresponding values $\{v_i\}$.
The retrieved features are integrated with the current representation and processed by subsequent model layers.

However, as explained before, directly applying such an attention-based retrieval in long-horizon imitation learning introduces two \emph{key limitations}.
First, because attention aggregates information from all stored history $\mathcal{M}_t$, the policy may attend to task-irrelevant details and incorporate them into decision-making, leading to spurious correlations between past observations and actions.
Second, during closed-loop execution, small prediction errors, compounded by interactions with the environment, accumulate over time.
These errors introduce noise into the stored representations, which can degrade latent representation quality, leading to model drift and cascading failures over long horizons.

\ourmethod{} addresses these limitations with two components (Fig.~\ref{fig:method_overview}).
First, we distill task-relevant priors from vision-language models (VLMs) into the policy via generated video question--answer (VQA) supervision.
This supervision encourages the retrieval to focus on task-relevant information, reducing reliance on irrelevant history (Sec.~\ref{sec:method:vqa-train},~\ref{sec:method:vqa-gen}).
Second, we restrict retrieval to the top-$k$ most relevant memory entries at each time step.
This sparsification limits the influence of accumulated error during closed-loop execution, enabling robust long-horizon control (Sec.~\ref{sec:method:topk}).

Concretely, \ourmethod{} trains a single visuomotor policy that jointly (i) imitates expert actions and (ii) answers VQA queries about the trajectory history.
Both objectives share the same modality-specific encoders $g_\theta$ and a common backbone $f_\theta$ that contains the memory retrieval mechanism, differing only in their prediction heads.
Top-$k$ sparsification is applied to retrieval for both objectives, encouraging the policy to condition only on a subset of relevant memory entries.
This shared training and policy weights encourage the retrieval mechanism to learn representations that are useful both for answering questions about the past and predicting actions.
As a result, supervision from VQA transfers to action prediction, guiding the policy to retrieve relevant information for the task.

\subsection{Distilling Vision-Language Model Priors via Video Question--Answering}\label{sec:method:vqa-train}

Vision-language models encode rich priors about scenes, objects, activities, and semantics to understand task instructions, making them a useful source of guidance for identifying task-relevant information in trajectories.
However, because VLMs are not grounded in action prediction, these priors alone do not capture all the information required for effective low-level control.
\ourmethod{} therefore distills priors from VLMs into the policy while grounding them through imitation learning.

\vspace{0.5em}
\noindent
\textbf{Video question--answering as retrieval supervision.}
In standard end-to-end imitation learning, the retrieval mechanism is trained solely through the action prediction objective.
\ourmethod{} additionally supervises the retrieval process using the video question--answering objective.
Specifically, we generate VQA pairs $(u, v)$, where $u$ is a text-based question probing task-relevant information from the trajectory history (\textit{e.g.}, object locations, object relations, event ordering, or subgoal progress), and $v$ is the corresponding answer.
For VQA supervision, the policy backbone is conditioned on the encoded history $\mathcal{M}_t$, the current observation embedding $x_t$, and the question $u$, and the answer is predicted via a VQA head:
\begin{equation}
    \hat{v} \sim \pi_\theta^{\text{vqa}}(v \mid \mathcal{M}_t, x_t, u),
\end{equation}
where $\pi_\theta^{\text{vqa}}$ consists of the shared observation $g_\theta^{\text{obs}}$, action encoder $g_\theta^{\text{act}}$, policy backbone $f_\theta$ (including the retrieval mechanism) with the action prediction objective, and a separate text prediction head $h_\theta^{text}$.

To answer $u$, the policy must retrieve the specific information from the history that supports the answer.
Since the questions are generated using vision-language models and tailored to the task, this objective injects task-relevant priors into the retrieval mechanism, providing direct supervision on what information should be retrieved.

\vspace{0.5em}
\noindent
\textbf{Joint grounding through imitation learning.}
While VQA supervision guides retrieval toward task-relevant information, it does not by itself ensure that the retrieved information is sufficient for low-level control.
We therefore jointly train the policy using an imitation learning objective:
\begin{equation}
    \mathcal{L}_{\text{IL}}(\theta)
    = \mathbb{E}_{(\tau_t, a_t^*) \sim \mathcal{D}}
    \left[-\log \pi_\theta^{\text{act}}(a_t^* \mid \tau_t, l)\right],
\end{equation}
where $a_t^*$ denotes the expert action.
We also define a VQA loss over the generated dataset $\widetilde{\mathcal{D}}$:
\begin{equation}
    \mathcal{L}_{\text{VQA}}(\theta)
    = \mathbb{E}_{(\tau_t, u, v) \sim \widetilde{\mathcal{D}}}
    \left[-\log \pi_\theta^{\text{vqa}}(v \mid \tau_t, u)\right].
\end{equation}
The final training objective is
\begin{equation}
    \mathcal{L}(\theta)
    = \mathcal{L}_{\text{IL}}(\theta)
    + \lambda \mathcal{L}_{\text{VQA}}(\theta),
\end{equation}
where $\lambda$ balances imitation and VQA supervision.
This joint training ensures that VQA provides informative priors about what information from the history should be retrieved, while imitation learning grounds retrieval in expert action prediction.

\subsection{Generating VQA Data from Long-Horizon Trajectories}\label{sec:method:vqa-gen}
\begin{figure}[htbp]
    \centering
    \vspace{-10pt}
    \includegraphics[width=0.5\textwidth]{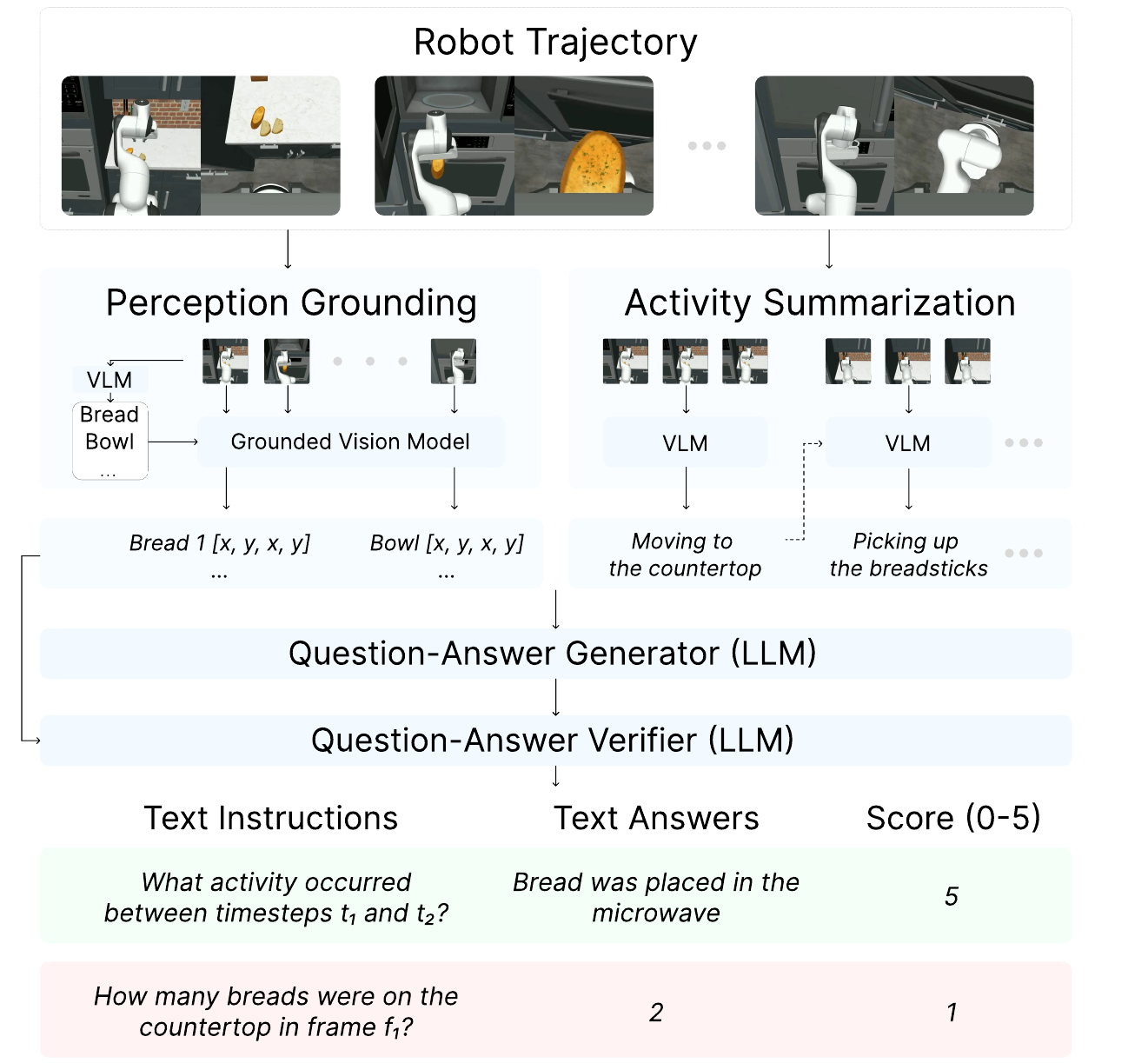}
    \vspace{-15pt}
    \caption{\textbf{Overview of video question-answer generation for knowledge distillation.} Images are first converted to text using a grounded vision model, and trajectories are summarized with each subtask and corresponding frame numbers. An LLM generates task-relevant question-answer pairs from these summaries and the task description, which are then verified for correctness and relevance, filtering out about $20\%$ of the data.}
    \label{fig:vqa_generation}
    \vspace{-15pt}
\end{figure}

Generating accurate VQA supervision for long-horizon robot trajectories is challenging. Single VLM calls over high-dimensional visual information often produce incorrect or hallucinated question--answer pairs, likely because state-of-the-art VLMs struggle to process long-context visual information coherently. To address this, we employ a multi-stage generation pipeline that produces high-quality VQA data (Fig.~\ref{fig:vqa_generation}).

\textit{(i) Trajectory-to-text summarization.}
Given a trajectory segment, we construct a textual description that captures scene elements (\textit{e.g.}, object identities and locations) and activity descriptions (\textit{e.g.}, ``the robot opened the drawer and placed the mug inside'').
To do so, we combine information from different specialized vision models, such as object detectors for spatial cues and VLMs for higher-level activity descriptions, and consolidate them into a single text representation of the trajectory.
This consolidation is performed only to generate questions and answers that provide retrieval priors; the information itself may not be sufficient to capture all the information required by the policy to predict low-level actions.

\textit{(ii) Question and answer synthesis.}
From the consolidated text, we synthesize questions that probe memory requirements relevant to the task.
We observe that language models tend to bias questions toward task-relevant information present in the initial portion of a trajectory.
To mitigate this bias, we explicitly prompt the language model to generate a question corresponding to a specific frame index, randomly sampled from the trajectory.
The model is also allowed to return `N/A' if no task-relevant information is present at the specified frame. We ask the language model to also generate the answer corresponding to the question.

\textit{(iii) Automatic filtering for correctness and relevance.}
Generated VQA pairs may contain incorrect or be weakly task-relevant or labeled as `N/A'.
To improve data quality, we rate each pair's correctness and task relevance using a language model. Pairs with low scores are filtered out.
The remaining set forms $\widetilde{\mathcal{D}}$, which is used for training.

See Sec.~\ref{sec:app:vqa_viz} for generated examples.

\begin{figure*}[t]
    \centering
    \includegraphics[width=0.98\textwidth]{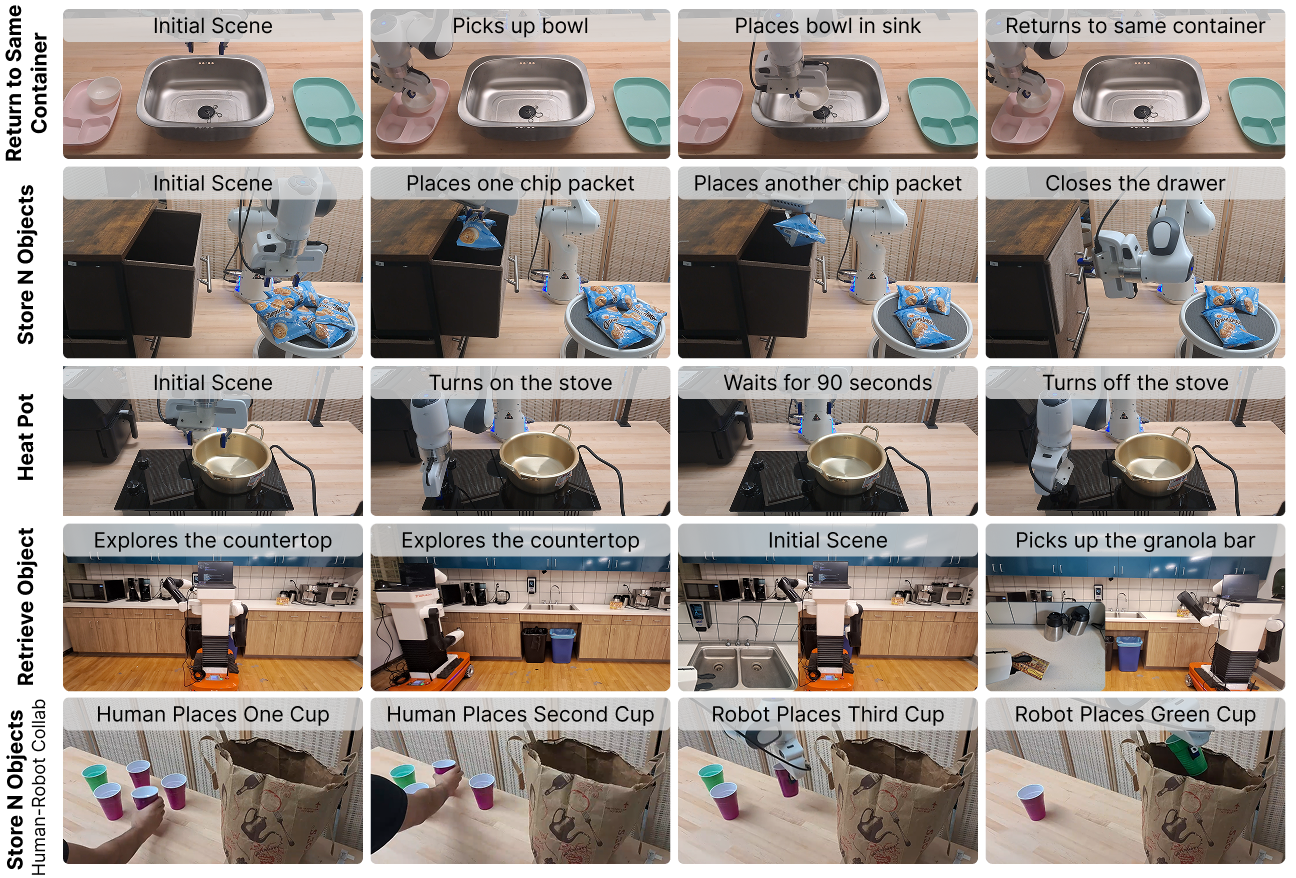}
    \vspace{-10pt}
    \caption{\textbf{Visualization of real-world tasks.} We evaluate \ourmethod{} in four stationary and mobile manipulation tasks, including a human-robot collaborative task (\textit{rows}) with strong partial observability, where the agent needs to retrieve different types of information from memory. The pictures in each column depict a different phase of the task.}
    \label{fig:rw_task_viz}
    \vspace{-10pt}
\end{figure*}

\subsection{Reducing Model Drift with Top-$k$ Attention Sparsification}
\label{sec:method:topk}
Even with improved retrieval supervision, long-horizon closed-loop execution remains susceptible to error accumulation.
In offline imitation learning, policies are trained on expert trajectories but must execute actions based on their own predictions at test time.
Small prediction errors can lead to future observations that deviate from the expected observations, leading to compounding errors over time.
Conditioning on large amounts of information from memory can further exacerbate this effect by repeatedly incorporating noisy information, resulting in cascading failures over long horizons.

To mitigate this issue, \ourmethod{} explicitly limits how much information from the history the policy can condition on at each time step.
Rather than attending to all available information, we first identify the information most relevant to the current decision and restrict retrieval to only these elements.
This allows the policy to focus on important information while reducing the effect of accumulated error in memory.

Concretely, let $\{m_i\}_{i=1}^{|\mathcal{M}_t|}$ denote candidate information extracted from the history, with keys $k_i = f_k(m_i)$ and values $v_i = f_v(m_i)$.
Given a query $q_t$, standard attention computes scores $s_i = \langle q_t, k_i \rangle$ and forms a weighted sum over all values.
In top-$k$ sparsification, we instead select
\begin{equation}
    \mathcal{I}_k = \mathrm{TopK}(\{s_i\}_{i=1}^{|\mathcal{M}_t|}, k),
\end{equation}
corresponding to the $k$ most relevant elements, and compute attention only over this subset with all elements outside $\mathcal{I}_k$ receive zero weight:
\begin{equation}
    c_t = \sum_{i \in \mathcal{I}_k}
    \frac{\exp(s_i)}{\sum_{j \in \mathcal{I}_k} \exp(s_j)} v_i.
\end{equation}

Since the top-$k$ operation is non-differentiable, we employ a straight-through estimator during training.
Intuitively, at each training step, the policy retrieves the top-$k$ elements based on the query-key similarity using the current model weights.
If the retrieved information improves action prediction or VQA accuracy, gradients flowing through the attention output increase the corresponding query-key similarities, reinforcing the selection of useful information.
Conversely, elements that do not contribute meaningfully receive weaker updates and are less likely to be selected in future steps.

To summarize, \ourmethod{} combines VQA-based distillation of informative VLM priors, which encourages task-relevant retrieval and reduces spurious correlations, with top-$k$ sparse attention, which limits error accumulation and mitigates model drift, yielding a policy that is effective for long-horizon, partially observable tasks.

\section{Experiments}
\subsection{Evaluation Tasks.}
We evaluate our approach on long-horizon manipulation tasks that require retrieving diverse information types---spatial, relational, numerical, and event-time---from past interactions, including human-robot collaborative scenarios.
These tasks span five scenarios reflecting diverse memory demands often faced in household settings (Fig.~\ref{fig:rw_task_viz}):
\\
\textbf{Retrieve Object} (\textit{Spatial information}).
This task requires recalling object locations observed earlier in the episode; the policy must fetch a granola bar from a location previously observed on the kitchen countertop, which is no longer visible at the time of action.
Across episodes, the object’s initial location varies between the left and right sides of the countertop.
\\
\textbf{Return to Same Container} (\textit{Object relations}).
This task requires recalling earlier relations among objects; after placing a bowl in a sink, the policy must later place it back on the plate from which it was originally taken.
Across episodes, the involved objects and their relations vary.
\\
\textbf{Store $N$ objects} (\textit{Numerical information}).
This task requires recalling a numerical count over time; the robot should place $N$ chip packets in a drawer before closing it, with the packets becoming occluded once placed.
Across episodes, both the initial number of chip packets and the target count $N$ vary.
\\
\textbf{Heat Stove} (\textit{Event-time information}).
This task requires recalling when an event occurred to act after a specified duration; the robot must heat a stove for a fixed duration.
Across episodes, the required heating duration varies between four, six, and eight minutes.
\\
\textbf{Store $N$ objects} (\textit{Spatial information, Human-robot collaboration}).
This task requires tracking a numerical count in a collaborative setting: a human places some purple cups first, then the robot completes the total of three purple cups in the shopping bag, and then adds a green cup to it.
Across episodes, the number of initial purple cups in the scene ranges from three to four, and the human places one to three cups.

\begin{table*}[t]
\centering
\caption{Evaluation on simulation benchmark with $50$ rollouts per task.}
\vspace{-8pt}
\small
\setlength{\tabcolsep}{7pt}
\renewcommand{\arraystretch}{1.2}
\begin{threeparttable}

\resizebox{\textwidth}{!}{%
\label{tab:sim_all}
\begin{tabular}{lccccc}
\toprule
\textbf{Method} 
& \textbf{Retrieve Object} 
& \textbf{Return to Same Container} 
& \textbf{Store N Objects} 
& \textbf{Heat Stove for $T$ mins.}
& \textbf{Average} \\

& \textit{(Spatial Info.)} 
& \textit{(Relational Info.)} 
& \textit{(Numerical Info.)} 
& \textit{(Event-time Info.)}
& \\
\midrule
Standard Transformer     
& $0.26$ & $0.23$ & $0.12$ & $0.27$ & $0.22$ \\
SAM2Act++                & $0.39$ & $0.11$ & $\mathbf{0.27}$ & $0.04$ & $0.20$ \\
ReMemBer                 & $0.36$ & $0.13$ & $0.21$  & $0.00$ & $0.18$ \\
Hand-Designed Features   & $\mathbf{0.68}$ & $0.15$ & $0.21$ & $0.13$ & $0.29$ \\
Scene Memory Transformer & $0.53$ & $0.25$ & $0.17$ & $\mathbf{0.40}$ & $0.34$ \\
Token Merging            &  $0.29$ &  $0.24$ & $0.25$ &  $0.38$ & $0.29$ \\
\midrule
HALO w/o VQA         & $0.42$ & $0.29$ & $0.17$ & $0.37$ & $0.31$ \\
HALO                 & $0.64$ & $\mathbf{0.32}$ & $0.26$ & $\mathbf{0.40}$ & $\mathbf{0.41}$ \\
\bottomrule
\end{tabular}%
}
\end{threeparttable}
\end{table*}

\begin{table*}[t]
\centering
\small
\setlength{\tabcolsep}{5pt}
\renewcommand{\arraystretch}{1.2}
\vspace{-8pt}
\caption{Evaluation on real-world tasks with $20$ rollouts per task}
\vspace{-16pt}
\begin{threeparttable}
\label{tab:rw_all}
\begin{tabular}{lcccccc}
\toprule
\textbf{Method} 
& \textbf{Retrieve Obj.}
& \textbf{Return to Container}
& \textbf{Store N Objs.}
& \textbf{Heat Stove}
& \textbf{Store N Objs.}
& \textbf{Average} \\
& 
& 
& 
& \textbf{for $T$ mins.}
& Human-Robot
& \\
& \textit{(Spatial Info.)}
& \textit{(Relational Info.)}
& \textit{(Numerical Info.)}
& \textit{(Event-time Info.)}
& \textit{Spatial Info.}
& \\
\midrule
Standard Transformer & $0.40$ & $0.30$ & $0.20$ & $0.40$ & $0.50$ & $0.36$ \\
HALO                 & $\mathbf{0.55}$ & $\mathbf{0.40}$ & $\mathbf{0.55}$ & $\mathbf{0.60}$ & $\mathbf{0.65}$ & $\mathbf{0.55}$ \\
\bottomrule
\end{tabular}
\end{threeparttable}
\vspace{-10pt}
\end{table*}

\subsection{Baselines.}
With the goal of reducing memory size or introducing inductive priors about what should be stored, prior work proposes several alternatives:
\\
\textbf{SAM2Act++}~\cite{sam2act} uses object-centric memory by storing object center pixel coordinates.
This induces a strong spatial prior and reduces noise in memory by storing only object locations.
\\
\textbf{ReMemBer}~\cite{anwar2025remembr} stores text summaries of past observations generated by a VLM.
This introduces semantic priors while compressing history into text.
\\
\textbf{Hand-designed Features} uses human-designed rules to select which task-relevant features to store or discard, similar to prior work MemER~\cite{memer}.
It encodes human priors about what information is important while keeping memory compact.
\\
\textbf{Scene Memory Transformer (SMT)}~\cite{fang2019scene} compresses the entire history into a single scene-level embedding via attention.
This reduces memory footprint but forces all past information into a fixed-size representation.
\\
\textbf{Token Merging}~\cite{memoryvla} compresses the history by merging tokens with similar embeddings that are temporally adjacent to maintain a fixed budget of memory.

\subsection{Results}
\textit{How well does \ourmethod{} perform compared to prior visuomotor imitation learning methods that incorporate alternative inductive priors?} (Table ~\ref{tab:sim_all})
\\
\textbf{Comparison to object-centric memory.}
Compared to SAM2Act++, \ourmethod{} improves average task success by $21\%$ points.
SAM2Act++ performs well when tasks align with its object-centric prior, such as counting objects, but struggles with tasks that require non-object-centric memory, particularly recalling event-time information.
This highlights the limitation of relying on a single type of prior.
\\
\textbf{Comparison to text-based summarization.}
\ourmethod{} outperforms ReMemBer by $23\%$ points on average.
Text summaries generated by VLM often capture coarse semantic information, but may omit details needed for control.
Textual summaries may capture object locations but omit action information for navigating to the object, explaining poor `Retrieve Object' performance.
It suggests that while VLM priors are useful, text-based compression alone is insufficient for control.
\\
\textbf{Comparison to hand-designed rule-based filtering.}
Compared to hand-designed features, \ourmethod{} achieves an absolute improvement of $12\%$.
While this performs on par with \ourmethod{} on spatial tasks, it remains less competitive than \ourmethod{} on other tasks.
This suggests that learning what to retrieve directly from data using~\ourmethod{} not only removes the need for manually designed task-specific priors but also improves performance, possibly by enabling the policy to exploit additional information for decision-making (App. Sec.~\ref{sec:app:retrieved_info_vis}).
\\
\textbf{Comparison to compressed memory representations.}
We compare \ourmethod{} against two approaches that compress history: Scene Memory Transformer (SMT), which uses a learned pooled attention layer to compress history into a single embedding, and Token Merging, which merges nearby tokens to maintain a fixed memory budget.
\ourmethod{} outperforms SMT by $7\%$ and Token Merging by $12\%$ points.
While compression reduces redundancies and noise, it can lose granular information necessary for low-level action prediction~\cite{shao2025tokens}.
In contrast, \ourmethod{} stores all information, enabling access to fine-grained details with selective retrieval to reduce the impact of noise.

Overall, we observe that no single method wins across all tasks. However,~\ourmethod{} remains competitive without task-specific assumptions or hand-designed rules, making it versatile with less engineering effort across information types. This versatility is reflected in average performance across tasks, alongside competitive performance on individual ones.
\\
\\
\textit{How effective is \ourmethod{} in real-world tasks?} (Table ~\ref{tab:rw_all})

We observe a similar trend in real-world settings, where \ourmethod{} consistently outperforms the standard Transformer baseline by $19\%$. Notably, this improvement persists on challenging tasks, including storing multiple objects, tasks completed by the human partner, and long-horizon tasks like heating a pot for up to eight minutes, which demand diverse types of information and long-context memory.

In addition, we measure manipulation and memory failures in real-world evaluations of the `Retrieve Object' task, finding that \ourmethod{} reduces them by $8\%$ and $25\%$ absolute over full attention, respectively. These results support our hypothesis that \ourmethod{} reduces model drift (fewer manipulation failures) and improves memory retrieval (fewer memory failures).
\\
\\
\begin{table}[t]
\centering
\caption{Evaluation on simulation with $50$ rollouts per task.}\label{tab:attn_type}
\vspace{-8pt}
\small
\setlength{\tabcolsep}{6pt}
\renewcommand{\arraystretch}{1.2}
\begin{threeparttable}
\resizebox{0.85\columnwidth}{!}{%
\begin{tabular}{lcc}
\toprule
\textbf{Method}
& \textbf{Retrieve Object} 
& \textbf{Return to Container} \\
\midrule
LSTM          & $0.14$ & $0.12$ \\
Mamba         & $0.20$ & $0.18$ \\
TransformerXL & $0.12$ & $0.20$ \\
\midrule
Window Attention   & $0.13$ & $0.16$ \\
Strided Attention & $0.20$ & $0.28$ \\
Hierarchical Attention & $0.28$ & $0.22$ \\
Gated Attention        & $0.45$ & $0.28$ \\
\midrule
HALO & $\mathbf{0.64}$ & $\mathbf{0.32}$ \\
\bottomrule
\end{tabular}
}
\end{threeparttable}
\vspace{-15pt}
\end{table}

\textit{What are the effects of VLM-induced priors and top-$k$ sparsification on performance?} (Table~\ref{tab:sim_all})
\\
\textbf{Effect of VLM-induced priors.}
We compare \ourmethod{} against a variant trained without VQA supervision. Adding VLM-induced priors improves average success by $10\%$, suggesting that VQA supervision helps the policy identify relevant history, particularly for tasks requiring recall of object placements or numerical information. Without it, the retrieval mechanism may attend to irrelevant history, degrading test-time performance.
\\
\textbf{Effect of top-$k$ sparsification.}
We compare top-$k$ retrieval to full attention over memory.
Top-$k$ improves performance by $9\%$.
Restricting retrieval reduces the amount of irrelevant or noisy information incorporated into decision-making, which is particularly important in long-horizon tasks where only a few past events are relevant to the current action.

See App. Sec.~\ref{sec:app:spurious_vis} and Sec.~\ref{sec:app:model_drift} for more analysis.
\\
\\
\textit{How does \ourmethod{} compare to alternative sparsification and memory mechanisms? (Table~\ref{tab:attn_type})}
\\
\textbf{Comparison to alternative sparsification methods.}
We compare top-$k$ sparsification to window attention~\cite{child2019generating}, strided attention~\cite{child2019generating}, hierarchical attention~\cite{beltagy2020longformer}, and gated attention~\cite{gatedattention}.
Fixed access patterns may miss task-relevant events occurring at unpredictable times; in contrast, top-$k$ selects entries based on content relevance, retrieving information tied to the task rather than position. Empirically, top-$k$ outperforms an alternative content-based sparsification method, gated attention by $11\%$ points.
\\
\textbf{Comparison to non-associative memory mechanisms.}
We compare against models that compress history into latent states, including LSTM~\cite{lstm} ($-20\%$), Mamba~\cite{mamba} ($-14\%$), and Transformer-XL~\cite{dai2019transformer} ($-12\%$).
While compressed-state models summarize past information efficiently, they may discard details needed for recalling specific past events.
Selective retrieval from explicit memory allows the policy to access such details when required.
\\
\\
\textit{What are the effects of different design choices in~\ourmethod{}?}
\vspace{0.5em}
\\
\textbf{Number of retrieved entries ($k$).}
We ablate the number of retrieved history entries ($k$) used by the attention mechanism. A moderate value ($k=8$) achieves the best performance ($52\%$ success). Smaller $k$ (\textit{e.g.}, $k=4$, $40\%$ success) omits useful context, while larger $k$ may incorporate irrelevant history ($48\%$ for $k=12$, $33\%$ for $k=16$, $44\%$ for $k=24$). Developing adaptive strategies that retrieve only the necessary amount of information at each step is a promising direction for future work.
\\
\textbf{Co-training vs.\ pretraining.}
We compare joint co-training of VQA and action prediction against a two-stage approach (VQA pretraining followed by action finetuning). Co-training VQA and action prediction achieves $64\%$ success, outperforming pretrain-then-finetune ($44\%$) and no-VQA training ($42\%$) by $20$ and $22$ points, respectively. The minimal gain from separate pretrain-then-finetune compared to no-VQA suggests that VQA knowledge is lost during fine-tuning, whereas co-training effectively shapes retrieval toward task-relevant information.
\\
\textbf{Stage-wise VQA generation vs. single video prompt generation.}
We compare \ourmethod{}'s stage-wise pipeline against a single-prompt baseline that generates QA pairs directly from long videos.
\ourmethod{} achieves $\mathbf{60\%}$ points more accurate QAs and $\mathbf{20\%}$ points fewer hallucinations.
The results suggest SOTA VLMs struggle with direct long-video reasoning for question--answer generation, highlighting the importance of the proposed stage-wise pipeline for generating high-quality VQA (Examples in App. Sec.~\ref{sec:app:vqa_viz}).

\section{Conclusion and Future Work}
In this work, we study offline imitation learning in partially observable domains where retrieving information from past observations and actions is essential to perform the task.
We present~\ourmethod{}, a method for training a visuomotor policy with a memory-retrieval mechanism for long-horizon imitation learning.
To address spurious correlations learned during training, \ourmethod{} co-trains action prediction with VLM-generated video question--answer supervision to bias retrieval toward task-relevant information.
To mitigate error accumulation from large memory, \ourmethod{} employs a top-$k$ sparsification strategy that restricts retrieval to the most relevant parts of the history.
Through comprehensive evaluation in simulation and real-world experiments across two robotic platforms, we demonstrate that our approach, which stores all information in memory and selectively retrieves relevant content, achieves significant improvements over alternative methods that reduce memory size through task-specific priors or learned compression.
We also find that VLM-induced priors are useful for selecting task-relevant information from memory, but, by themselves, can be insufficient for low-level control. 
In the future, we plan to investigate mechanisms to switch from fixed to active querying, adapting to each task.
Moreover, we will explore broadening the scope of querying to cross-episodic information, using memory to learn from failures.
Additionally, because retrieval from large memory adds retrieval latency, especially as we expand the context to longer time periods, we will explore speculative retrieval methods to hide it.

\section{Acknowledgments}
We thank Huihan Liu, Dvij Kalaria, and Yifeng Zhu for providing valuable feedback on the manuscript.
We also thank all members of the Robot Perception and Learning
Lab and the Robot Interactive Intelligence Lab at UT Austin for their insightful discussions.
This work was partially supported by the National Science Foundation (FRR-2145283, EFRI-2318065), the Office of Naval Research (N0001424-1-2550), the DARPA TIAMAT program (HR0011-24-90428), and the Army Research Lab (W911NF-25-1-0065).
It was also supported by the Institute of Information \& Communications Technology Planning \& Evaluation (IITP) grant funded by the Korean Government (MSIT) (No. RS2024-00457882, National AI Research Lab Project).

\bibliographystyle{plainnat}
\bibliography{references/neuroscience, references/ai, references/appendix}
\clearpage
\section{Appendix}
\begin{figure*}[h]
    \centering
    \includegraphics[width=\textwidth]{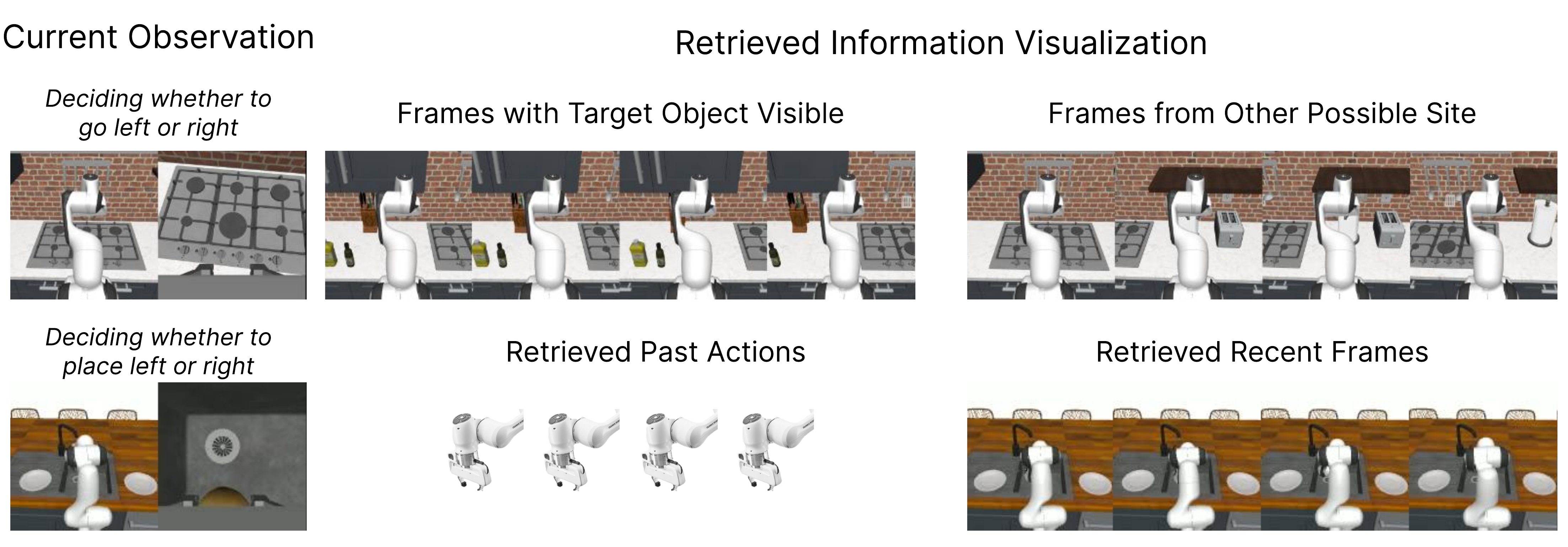}
    \caption{Qualitative visualization of retrieved memory for two decision points. Left: current observations where the robot must choose a movement direction based on past information. Right: retrieved frames and action embeddings used for decision-making.}
    \label{fig:retrieval-viz}
    \vspace{-0.5em}
\end{figure*}

\subsection{Qualitative Analysis of Retrieved Information}\label{sec:app:retrieved_info_vis}
Qualitative examples are shown in Fig.~\ref{fig:retrieval-viz}. In the top example (Retrieve Object), the robot must choose a navigation direction based on where the target object was observed earlier in the episode. In the bottom example (Return to Same Container), the robot must place an object back into the correct container based on which container it was picked up from previously.

We notice that the policy retrieves heterogeneous and complementary information: for Retrieve Object (top, navigation), it retrieves both, frames where the target object (olive oil) was visible and frames from other locations where the target object is usually found but not currently placed; for Return to Same Container (bottom, manipulation), it retrieves both past actions potentially indicating the object's original container and past frames indicating subgoal progress (`picked-up' vs. `placed').

This behavior goes beyond what simple, human-designed memory features (\textit{e.g.}, storing only frames where the object is visible) may capture. Together with the quantitative gains over hand-designed features  (Table~\ref{tab:sim_all}), these qualitative results suggest that learning what to retrieve directly from data with \ourmethod{} is more effective than manually specifying what to store, particularly for long-horizon, partially observable tasks.

\subsection{Qualitative Visualization of Generated VQA Data}\label{sec:app:vqa_viz}
\begin{figure*}[h]
    \centering
    \includegraphics[width=\textwidth]{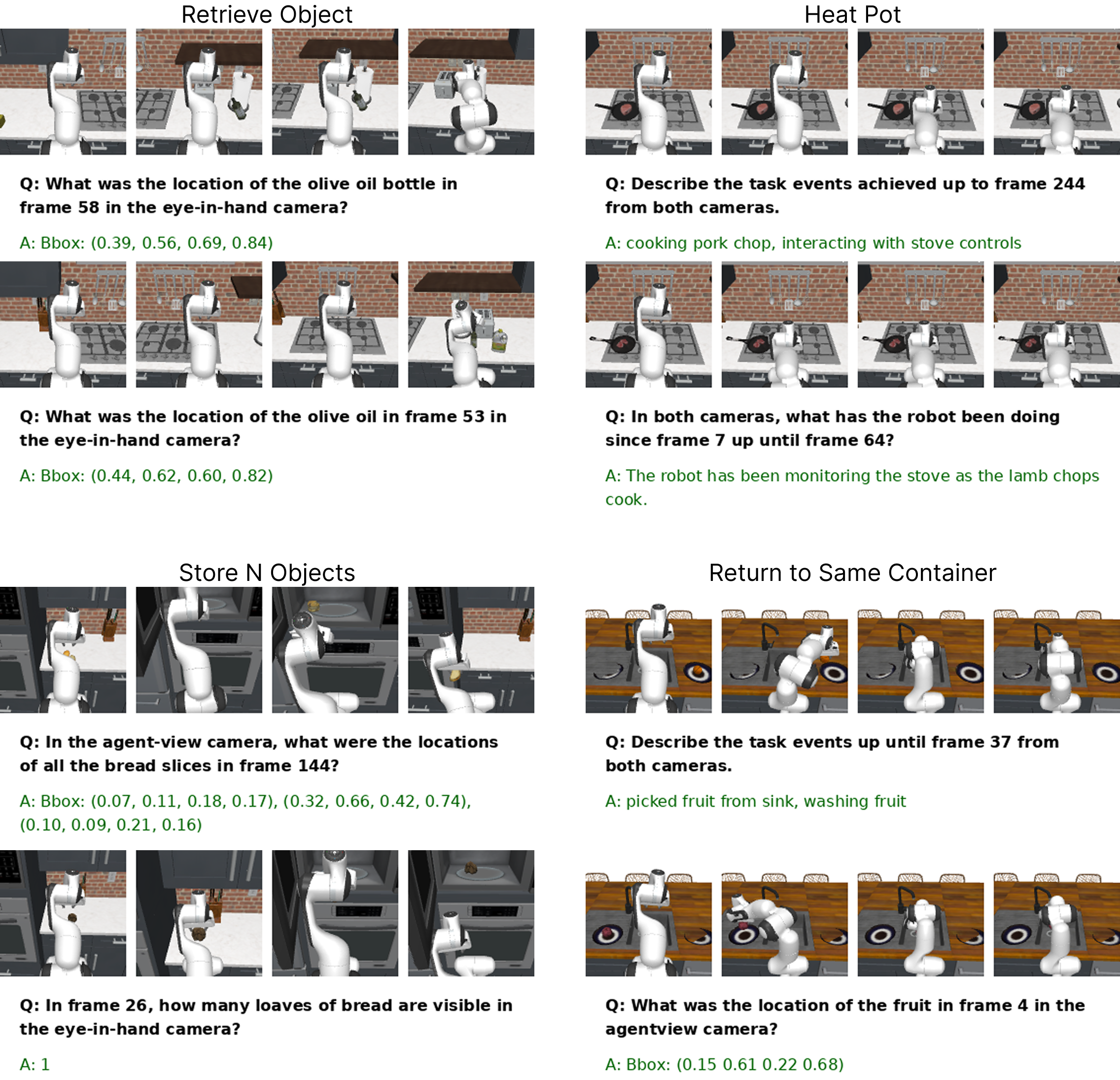}
    \caption{Examples of automatically generated video question--answer pairs across tasks, illustrating different types of task-relevant information required to answer the queries. We only show four frames uniformly sampled from the video to maintain visual clarity.}
    \label{fig:qa_examples}
    \vspace{-0.5em}
\end{figure*}

We visualize examples of the generated VQA supervision used to train retrieval in Fig.~\ref{fig:qa_examples}. 
The questions probe different types of information, \textit{e.g.}, object locations, counts, and long-horizon task progress. 
These examples demonstrate that the generated VQA data provides diverse, task-relevant supervision for guiding memory retrieval. 
\subsection{Qualitative Visualization of Spurious Correlations}\label{sec:app:spurious_vis}
\begin{figure*}[t]
    \centering
    \includegraphics[width=0.98\textwidth]{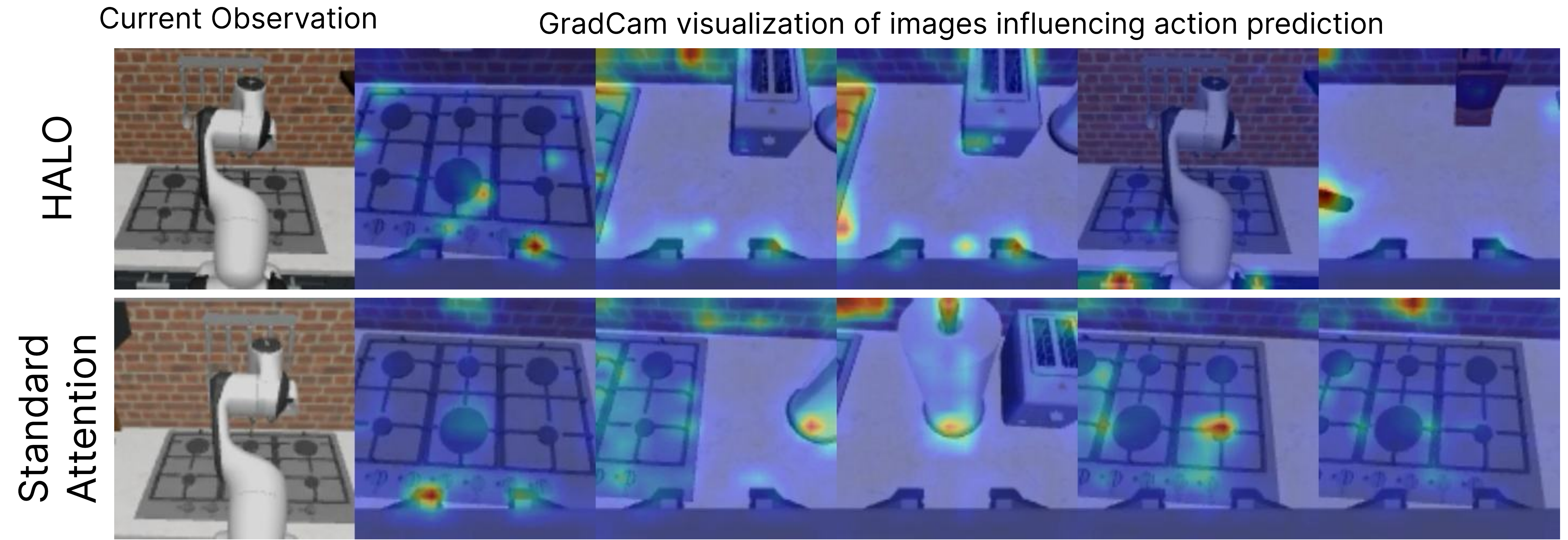}
    \caption{Grad-CAM-style visualizations of image regions from history influencing action prediction. \textbf{Top:} \ourmethod{} focuses on the target object location (top-rightmost). \textbf{Bottom:} Without VQA supervision (standard Transformer), distractors may spuriously influence the action prediction (\textit{e.g.}, tissue roll position).}
    \label{fig:gradcam_viz}
\end{figure*}
We provide qualitative visualizations of image regions from the observation history that most influence the action prediction, following a GradCAM-style method~\cite{gradcam}.
A comparison between \ourmethod{} and the standard Transformer offers qualitative insight into spurious correlations that may influence action prediction in the absence of VQA supervision.

As shown in Fig.~\ref{fig:gradcam_viz}, \ourmethod{} focuses more strongly on task-relevant target objects (\textit{e.g.}, the olive oil bottle), whereas the standard Transformer exhibits sensitivity to irrelevant distractors (\textit{e.g.}, the tissue roll position).
Together with the quantitative results in Table~\ref{tab:rw_all}, these visualizations suggest that adding VQA supervision reduces reliance on spurious visual correlations.
\subsection{Quantitative Analysis of Model Drift Errors}\label{sec:app:model_drift}
{
\centering
\includegraphics[width=0.80\columnwidth]{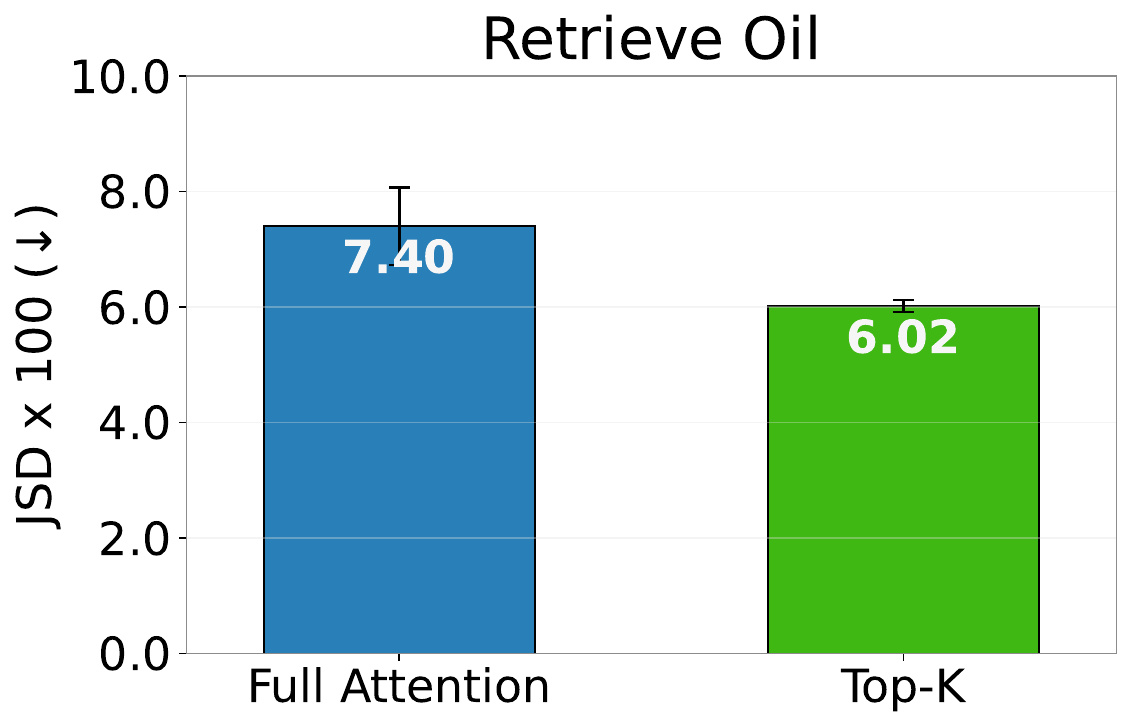}
\label{fig:jsd_action}
}

\\
To quantify drift from expert behavior, we measure the distance between the policy's action distribution during rollouts and the expert's action distribution from teleoperated demonstrations. Both rollouts and expert trajectories are evaluated from identical initial environment configurations, using privileged resets in simulation to isolate differences attributable to the learned policy rather than to initial-state variation.

We quantify action distribution drift using the Jensen-Shannon divergence (JSD), a symmetric and bounded measure of discrepancy between probability distributions. For each task, we aggregate actions from rollouts and the corresponding expert demonstrations. Since actions are continuous, we discretize each action dimension into fixed bins and construct normalized histograms over the joint action space. We then compute the total JSD between the flattened rollout and expert histograms, yielding a single scalar per task that summarizes overall policy drift: larger values indicate greater deviation from expert behavior, while smaller values indicate closer alignment.

Using this metric, we find that incorporating top-$k$ attention reduces the JSD between rollout and expert action distributions compared to full attention ($0.06$ vs. $0.07$), indicating reduced model drift and possibly contributing to a higher success rate as seen in Table~\ref{tab:sim_all} (`\ourmethod{} w/o VQA' vs. `Standard Transformer').
\\
\noindent
\textbf{Note.} We emphasize that JSD is a diagnostic metric and does not directly determine task success; exact matching of expert actions may be suboptimal under environment stochasticity. Nevertheless, reduced JSD provides direct evidence that top-$k$ sparsification mitigates reduced drift, complementing the primary performance gains in Table~\ref{tab:sim_all}.

\subsection{Quantitative analysis of the effect of $k$ with context length $N$}
\begin{figure}[h]
    \centering
    \vspace{-0.35cm}
    \includegraphics[width=\linewidth]{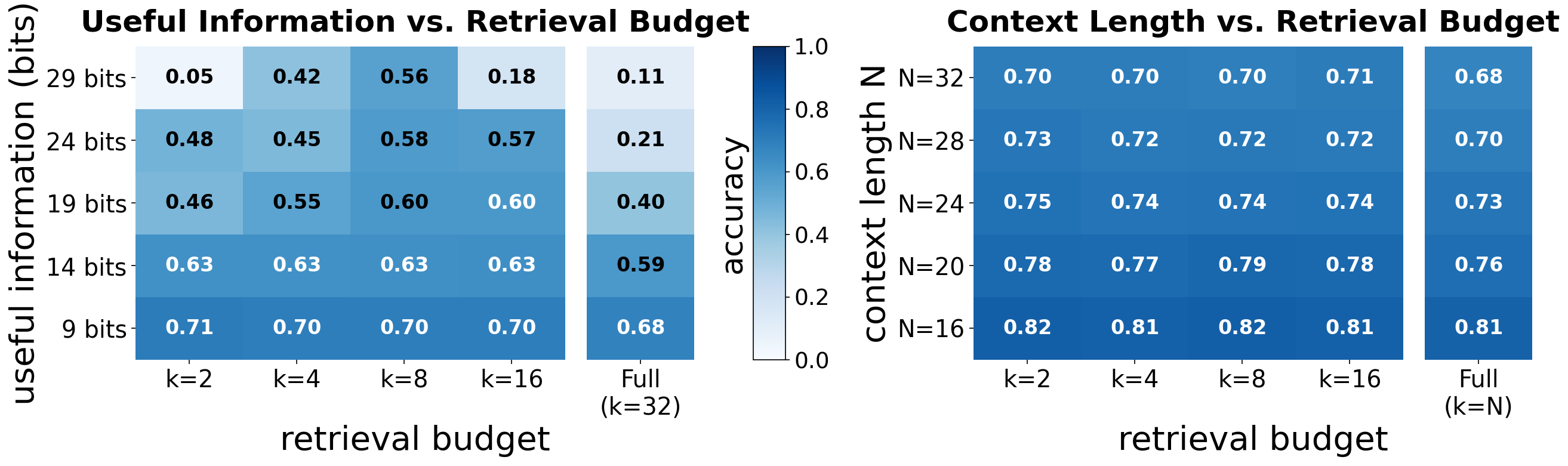}
    \vspace{-0.84cm}
    \label{fig:k_ablation}
\end{figure}
We hypothesize that the retrieval budget $k$ should depend on the information required for a decision, rather than on the context length $N$. While isolating this in manipulation is non-trivial, we evaluate on a controlled toy dataset by varying the minimum bits of information required to make a prediction, and the context length $N$. We observe that for fixed $N$, performance is similar across $k$ values but varies across tasks with different information requirements.

\subsection{Implementation Details}\label{sec:impl}
\noindent
\textbf{Architecture.}
The observation encoder $g_\theta^{\text{obs}}$ consists of a pretrained, frozen vision model~\cite{crossmae} followed by a learned pooling layer that maps sensory inputs (\textit{e.g.}, RGB and proprioception) to a single embedding.
The action encoder $g_\theta^{\text{act}}$ embeds past actions using an MLP.
The policy backbone $f_\theta$ is a Transformer with four layers of local attention with four heads each and two layers that perform long-horizon retrieval using top-$k$ attention.
The action head $h_\theta^{\text{act}}$ is an MLP that predicts $32$ future actions, and the text head $h_\theta^{\text{text}}$ is a single Transformer block that autoregressively predicts answers over the standard Qwen3 vocabulary~\cite{qwen3}.
\\
\textbf{Top-$k$ attention.}
We maintain an explicit memory buffer $\mathcal{M}_t$ containing $192$--$3840$ entries per episode, depending on the maximum task horizon.
Entries are added to the memory every eight interaction steps, and each entry stores either per-camera observation embeddings $x_i$ or past action embeddings $e_i$.
At each time step, top-$k$ attention selects $k=8$ entries based on cosine similarity.
We use a straight-through estimator to enable backpropagation through the top-$k$ selection.
\begin{table*}[h]
\centering
\caption{Task settings across simulation and real-world domains.}
\label{tab:task-settings}
\resizebox{0.75\textwidth}{!}{%
\begin{tabular}{lllll}
\toprule
\multirow{2}{*}{\textbf{Task Name}} 
& \multirow{2}{*}{\textbf{Domain}} 
& \multirow{2}{*}{\shortstack{\textbf{Max Memory} \\ \textbf{Storage}}}
& \multirow{2}{*}{\shortstack{\textbf{\# Cameras}}}
& \multirow{2}{*}{\shortstack{\textbf{\# Expert} \\ \textbf{Demos}}} \\
\\
\midrule
Retrieve Object & Simulation & $196$ & $2$ & $50$ \\
Retrieve Object & Real-world & $192$ & $1$ & $50$ \\
Return to Same Container & Simulation & $192$ & $2$ & $50$ \\
Return to Same Container & Real-world & $192$ & $2$ & $100$ \\
Store $N$ Objects & Simulation & $768$ & $2$ & $50$ \\
Store $N$ Objects & Real-world & $768$ & $2$ & $100$ \\
Heat Stove for $T$ mins & Simulation & $768$ & $2$ & $50$ \\
Heat Stove for $T$ mins & Real-world & $768$ & $2$ & $60$ \\
\bottomrule
\end{tabular}
}
\end{table*}

\\
\noindent
\textbf{Losses and weighting.}
The overall training objective is $\mathcal{L} = \mathcal{L}_{\text{IL}} + \lambda \mathcal{L}_{\text{VQA}}$, with $\lambda=\{0.1,1.0\}$.
For imitation learning, actions are optimized using an $\ell_1$ loss.
For VQA, answers are optimized using a cross-entropy loss.
\\
\textbf{VQA generation and filtering.}
Trajectory-to-text summaries are generated using SAM3~\cite{sam3} and gpt-4o-mini~\cite{gpt-4o}.
For each trajectory, we randomly sample frame indices to ask a particular question for and prompt GPT-4o-mini to produce corresponding $(u, v)$ pairs.
Generated pairs are filtered using gpt-4o~\cite{gpt-4o} as a judge model based on correctness and relevance. Both the prompts for generation and verification are provided in Sec.~\ref{sec:app:prompts}.
The final VQA dataset contains $20$k--$30$k question--answer pairs per task.
\\
\begin{table*}[h]
\centering
\caption{Training hyperparameters.}
\label{tab:training-config}
\resizebox{\columnwidth}{!}{%
\begin{tabular}{ll}
\toprule
\textbf{Config} & \textbf{Value} \\
\midrule
Optimizer & AdamW \\
Base learning rate & $5\times10^{-4}$ \\
Effective batch size & $32$ \\
Weight decay & $0.01$ \\
Warmup epochs & $2$ \\
Total epochs & $200$ \\
Early stopping epochs & $50$ \\
Action prediction horizon & $32$ \\
Proprioception noise (std) & $0.005$ \\
Brightness augmentation & $\mathrm{Uniform}(-0.1,\,0.1)$ \\
Contrast augmentation & $\mathrm{Uniform}(0.8,\,1.2)$ \\
\bottomrule
\end{tabular}
}
\end{table*}

\textbf{Training details.}
All models are trained end-to-end using the hyperparameters listed in Table~\ref{tab:training-config}, with separate models for each task.
\\
\textbf{Evaluation protocol.}
Policies are evaluated in a closed loop over $50$ episodes per task in simulation and $20$ episodes per task in the real world.
We report the final task success rate.

\subsection{Prompts}\label{sec:app:prompts}
Please see the bottom of the appendix.
\begin{FullWidthPromptBox}{Question--Answer Generation Prompt}
You are given a task instruction, task's description, summary of task events labeled by frame numbers, and a list of information provided per-frame and per-camera about the visible objects in the scene, their locations, and robot actions.
You are given a separate list of QUERY-FRAME INDEXES to generate the query and answer.

Your job:
1. Identify events relevant to the task.
  Use the task description, and summary of task events to understand the scene in the current episode.
  Using these information, describe what plausibly happens in the task for each of the given QUERY-FRAME INDEXES.
2. Write ONE memory-based query important for the task for each of the given QUERY-FRAME INDEXES
  The question must require recalling earlier information provided in the prompt for each of the given QUERY-FRAME INDEXES. It may include: 
    - object location seen earlier in the QUERY-FRAME INDEX (e.g., bounding box coordinates, etc.)
    - number of objects of a particular type seen earlier in the QUERY-FRAME INDEX (e.g., number of sponges, etc.)
    - few words about the task events achieved upto the QUERY-FRAME INDEX using the provided summary of task events (e.g., picked two sponges, placed one in sink)
    - time-frame of the task events which includes the QUERY-FRAME INDEX (e.g., frame 10 to frame 20)
    - important object relations to remember (e.g., object X is to the left of object Y)
  Rules (VERY IMPORTANT):
    - The answer should be relevant to the provided task instruction and help in solving the task.
    - The questions should be about the past events or observations upto the QUERY-FRAME INDEX, not the future events.
    - Include when, what, where, how much, how many, description, etc. in the question, especially depending on the task demands.
    - Questions should NOT include generic references like 'first frame,' 'last frame' but rather use the QUERY-FRAME INDEX provided in the prompt to indicate the frame number in the question.
    - Do NOT include questions not used or important for completing the task.
  Example 1 (with QUERY-FRAME INDEX: 19), Query: "In the eye-in-hand camera, how many sponges were there in frame 19?"
  Example 2 (with QUERY-FRAME INDEX: 12), Query: "What was the location of the blue sponge in frame 12 in the agent-view camera?"
  Example 3 (with QUERY-FRAME INDEX: 38), Query: "Describe the task events up until frame 38 from both cameras?"
  Example 4 (with QUERY-FRAME INDEX: 24), Query: "What was the location all the sponges in frame 24 in the agent-view camera?"
  Example 5 (with QUERY-FRAME INDEX: 13, between frame 10 and frame 20), Query: "When did the robot pick up the blue sponge?"
3. Answer the query
  Provide the single best answer using only the provided information provided for each of the given QUERY-FRAME INDEXES and your event notes.
  The answer should be appropriate for the referenced QUERY-FRAME INDEX and the camera. If there's no answer, provide N/A
  The answer should be concise using very few words. Make sure if there are multiple answers to the query, provide all the answers.
  Example 1, 2 sponges
  Example 2, Bbox: (x_min, y_min, x_max, y_max) (if there are multiple bboxes, provide all the bboxes separated by a commas)
  Example 3, picked one sponge, placed one in trash, picked another sponge
  Example 4, Bbox: (x_min, y_min, x_max, y_max), (x_min, y_min, x_max, y_max), ...
  Example 5, Frame: 10 to frame 20

OUTPUT FORMAT:
The JSON should contain a single "results" field with a list of dictionaries, each containing the three required fields for each of the specified QUERY-FRAME INDEX:
json{{
  "results": [
    {{
      "description-of-query": "counting / task-events / location / etc, camera-name",
      "query": "<one memory-based question for the first QUERY-FRAME INDEX>",
      "answer": "<concise answer grounded in the frames for the first QUERY-FRAME INDEX>"
    }},
    ...
}}

TASK INSTRUCTION: \textit{task_language}

TASK DESCRIPTION: \textit{task_description}
\end{FullWidthPromptBox}

\begin{FullWidthPromptBox}{Question-Answer Filtering Prompt}
You are evaluating multiple candidate outputs for a single robot episode.
You will be given the SAME EPISODE CONTEXT as in the base prompt:
- A task instruction, task's detailed description, summary of task events labeled by frame numbers, and a list of information provided per-frame and per-camera about the visible objects in the scene, their locations, robot actions, etc.
After the task information, you will also be given a BUNDLE of candidate items to evaluate.
Each candidate item includes:
  - id (MUST be an integer)
  - 'query' (a memory-based question),
  - 'answer' (the model's answer),

Your job (per item):
1) Task alignment: Do query, instruction, and answer all help in solving the task as described in the task instruction and task description?
2) Grounding: Is the answer specific and consistent with the episode's referenced frame(s) indices and camera(s)? Is the answer correct? Verify for both.
3) Memory dependence: Does the query REQUIRE recalling earlier information about the visible objects in the scene, their locations, robot actions, etc.
4) Instruction quality: Does the instruction briefly prime what to remember?

Answer should be correct for high scores 3--5, both inclusive.
Scoring (integers only, 1--5):
5 = Excellent: query and answer clearly help in solving the task, grounded and answers are correct, memory-dependent question
4 = Good: query and answer help in solving the task; grounding and answer are correct; partially memory-dependent question
3 = Fair: query and answer help in solving the task; ground and answer are correct; not memory-dependent question
2 = Weak: the question-answer pair is not useful to remember for the task (e.g., asking about distractor objects, not relevant to the task); answer is correct;
1 = Poor: irrelevant to the task; answer is incorrect;
Decision rule:
- If score >= 3 \rightarrow decision = ``keep''
- If score <= 2 \rightarrow decision = ``skip''

INPUT FORMAT
You will receive:
1) The task information: a task instruction, task's detailed description, and a list of information provided per-frame and per-camera about the visible objects in the scene, their locations, robot actions, etc.
2) Then a bundle of candidate items:
[
    {{
      "id": "<integer>",
      "query": "<string>",
      "answer": "<string>",
    }},
    ...
]

STRICT OUTPUT FORMAT (no prose, no extra fields in the JSON). Mention one entry per ID for each entry in the candidate items. Do NOT skip any IDs:
json{{
  "results": [
    {{"reasoning": "<explain briefly why it will be useful to remember for the task>", "id": <integer>, "score": <integer>, "decision": "keep"|"skip"}},
    ...
  ]
}}
Important:
- 'id' in output MUST match the integer 'id' from input.
- Base your judgments ONLY on the provided task information (task instruction, task description, and provided information) and the candidate fields.
- Do NOT include explanations, intermediate notes, or any fields other than id, score, and decision in the JSON.

TASK INSTRUCTION: {task_language}

TASK DESCRIPTION: {task_description}
\end{FullWidthPromptBox}

\end{document}